\documentclass{article}
\usepackage[greek.polutoniko,english]{babel}
\usepackage[utf8]{inputenc} 
\usepackage[a4paper, total={6in,8in}]{geometry}
\usepackage[affil-it]{authblk}
\usepackage{amsmath}
\usepackage{mathtools}
\usepackage{graphicx}
\usepackage{subcaption}
\usepackage{booktabs}
\usepackage{setspace}

\large
\begin{document}
\title{Enhancing Cross Entropy with a Linearly Adaptive Loss Function for Optimized Classification Performance}

\author[1,2,3,*]{Jae Wan Shim}

\affil[1]{Extreme Materials Research Center, Korea Institute of Science and Technology, 5 Hwarang-ro 14-gil, Seongbuk, Seoul, 02792, Republic of Korea}
\affil[2]{Climate and Environmental Research Institute, Korea Institute of Science and Technology, 5 Hwarang-ro 14-gil, Seongbuk, Seoul, 02792, Republic of Korea}
\affil[3]{Division of AI-Robotics, KIST Campus, University of Science and Technology, 5 Hwarang-ro 14-gil, Seongbuk, Seoul, 02792, Republic of Korea}

\affil[*]{jae-wan.shim@kist.re.kr}
\maketitle

\begin{abstract}
We propose the Linearly Adaptive Cross Entropy Loss function. This is a novel measure derived from the information theory. In comparison to the standard cross entropy loss function, the proposed one has an additional term that depends on the predicted probability of the true class. This feature serves to enhance the optimization process in classification tasks involving one-hot encoded class labels. The proposed one has been evaluated on a ResNet-based model using the CIFAR-100 dataset. Preliminary results show that the proposed one consistently outperforms the standard cross entropy loss function in terms of classification accuracy. Moreover, the proposed one maintains simplicity, achieving practically the same efficiency to the traditional cross entropy loss. These findings suggest that our approach could broaden the scope for future research into loss function design.
\end{abstract}

\flushbottom

\thispagestyle{empty}


\section*{Introduction}

\label{sec:intro}
One of loss functions using in machine learning, cross entropy has its roots in information theory \cite{shannon1948, shannon1948b}. The concept of cross entropy is closely related to the relative entropy that is also known as Kullback-Leibler divergence \cite{Kullback}. 

In the original article of Kullback and Leibler, the term of ``divergence" was used for a symmetric formula in contrast to the modern usage of asymmetric one. Note that the symmetric Kullback-Leibler divergence was already introduced by Jeffreys \cite{Jeffreys} and interpreted by a measure of the discrepancy between two laws of chance, i.e. probabilities. He stated invariance for non-singular transformations in addition to the symmetry, positive definiteness, and additivity.

Cross entropy, which is an asymmetric quantity, is a measure of the dissimilarity between two probability distributions, typically one true distribution and one estimated distribution and indicates how well the estimated distribution approximates the true distribution. In the context of machine learning, it is often used as a loss function for classification tasks, comparing the predicted probability distribution to the true distribution \cite{goodfellow}. Given two discrete probability distributions $P$ and $Q$, the cross entropy $H(P, Q)$ can be calculated using the following formula:

\begin{equation}
H(P, Q) = - \sum_{i} P(x_i) \log Q(x_i)
\end{equation}

Here, $P(x_i)$ is the probability of event $x_i$ according to distribution $P$, and $Q(x_i)$ is the probability of event $x_i$ according to distribution $Q$. In the case of neural networks for classification tasks, $P$ typically represents the true probability distribution, and $Q$ represents the predicted probability distribution generated by the model.

One of the early examples of the form of cross entropy being used in machine learning can be found in the work of Nelder and Wedderburn \cite{Nelder}. They introduced the Generalized Linear Model (GLM) framework in their paper. Although GLMs are not explicitly neural networks, the framework includes the concept of deviance for a binomial distribution, which is closely related to cross entropy with additional terms \cite{mccullagh}.

Recent research on loss functions has introduced several approaches: A theoretical analysis of a broad family of loss functions, comp-sum losses, is explored in Mao et al. \cite{mao}. To address the limitation of softmax loss, which cannot guarantee that the minimum positive sample-to-class similarity exceeds the maximum negative sample-to-class similarity, Zhou et al. \cite{Zhou} proposed the Unified Cross-Entropy. Li et al. \cite{Li} proposed an approximation of the gradients of cross-entropy loss to avoid the vanishing gradient problem of the cross-entropy loss. In a different approach, Bruch \cite{Bruch} developed a cross entropy-based learning-to-rank loss function, and Rezaei-Dastjerdehei et al. \cite{Rezaei} suggested weighted binary cross-entropy to solve imbalance problem. Further elaborations and analyses of cross entropy can be found in Zhang and Sabuncu \cite{Zhang2018}, Martinez and Stiefelhagen \cite{Martinez}, Li et al. \cite{li2019dual}, Pang et al. \cite{pang2019rethinking}, Bruch et al. \cite{bruch2019analysis}, Zhou et al. \cite{zhou2019mpce}, and Ho and Wookey \cite{Ho}.

When dealing with one-hot encoded class labels in classification tasks, the true probability distribution $P$ is represented as a one-hot vector, where only one element is equal to 1 (corresponding to the true class) and all other elements are 0. Let's denote the number of classes as $C$ and the true class label as $c$. The one-hot encoded vector for class $c$ can be represented as $P(x_i) =  1$  if  $i = c$, and $0$ otherwise. The predicted probability distribution $Q$, generated by the model (e.g., the softmax output), is a vector of probabilities $[q_1, q_2, \dots, q_C]$, where each $q_i$ corresponds to the predicted probability of class $i$. Given this, the cross entropy loss $H(P, Q)$ for one-hot encoded class labels can be simplified as follows:
\begin{equation}
\label{eq:crossEntropyOneHot}
H(P, Q) = - \log Q(x_c)
\end{equation}
In this case, the cross entropy loss is the negative logarithm of the predicted probability for the true class. Minimizing this loss function encourages the model to increase the predicted probability of the true class, which in turn improves classification performance. The learning process indeed focuses on optimizing the predicted probability for the true class while not directly considering the information from the false classes. While it's true that the model does not explicitly use the information from the false classes during the learning process, by increasing the predicted probability of the true class, the model is implicitly reducing the predicted probabilities of the false classes.

Note that for one-hot encoded class labels, we obtain the same Eq.~\ref{eq:crossEntropyOneHot} as the cross entropy loss from Kullback-Leibler divergence,
\begin{equation}
D(P, Q) = - \sum_{i} P(x_i) \log \frac {Q(x_i)}{P(x_i)}.
\end{equation}

In this article, the ``Linearly Adaptive Cross Entropy Loss function" is proposed, which is a novel loss function designed to enhance the optimization process in machine learning classification tasks.  We provide a theoretical foundation by deriving the linearly adaptive cross entropy loss function based on foundational concepts in information theory. This function has a unique feature that is the inclusion of an additional term that depends on the predicted probability of the true class.

We present compelling evidence that the linearly adaptive cross entropy loss function consistently outperforms the conventional cross entropy loss function throughout the training process. The proposed function increases computation complexity only by two operations of one subtraction and one multiplication, achieving practically the same efficiency to the traditional cross entropy loss. 

Through this theoretical derivation and extensive empirical validation, we provide substantial evidence in support of our claim that the linearly adaptive cross entropy loss function can improve the optimization process and enhance classification accuracy in machine learning tasks. Further details, including specific results and visual illustrations, will be presented in the following sections.

\section*{Linearly Adaptive Cross Entropy Loss}
Here we suggest a linearly adaptive cross entropy loss as the following:
\begin{equation}
Adp(P, Q) = - \left[1-Q(x_c)\right]\log Q(x_c). 
\end{equation}
The proposed loss function combines aspects of the cross entropy loss with an additional term that depends on the predicted probability of the true class. 
The above proposed loss can be obtained from Jeffreys divergence $J(P, Q)$ or the symmetric Kullback-Leibler divergence:
\begin{equation} \label{eq:jeffrey}
J(P, Q) = D(P, Q) + D(Q, P)
\end{equation}
or
\begin{equation}
J(P, Q) =  - \sum_{i} P(x_i) \log \frac {Q(x_i)}{P(x_i)} - \sum_{i} Q(x_i) \log \frac {P(x_i)}{Q(x_i)}.
\end{equation}
For one-hot encoded class labels, we can simplify $D(P, Q)$ and $D(Q, P)$ as
\begin{equation} \label{eq:simple_DPQ}
D(P, Q) \approx - \log Q(x_c)
\end{equation}
and
\begin{equation} \label{eq:simple_DQP}
D(Q, P) \approx Q(x_c) \log Q(x_c).
\end{equation}
To clarify the above simplification in Eq.~(\ref{eq:simple_DPQ}), we address that the one-hot encoded vector for class $c$ can be represented as $P(x_i) =  1$  if  $i = c$, and $0$ otherwise. It is considerable to use a very small value instead of zero to avoid singularities such as division by zero or multiplication by infinity. Also, Eq.~(\ref{eq:simple_DQP}) is obtained by dropping the false classes terms  in $D(Q, P)$. Note that the predicted probability distribution $Q$, generated by the model (e.g., the softmax output), is a vector of probabilities $[q_1, q_2, \dots, q_C]$, where each $q_i$ corresponds to the predicted probability of class $i$.  We finally obtain the linearly adaptive cross entropy loss $Adp(P, Q)$ from Eq.~(\ref{eq:jeffrey}) with the above simplified $D(P, Q)$ and $D(Q, P)$, which are Eq.~(\ref{eq:simple_DPQ}) and Eq.~(\ref{eq:simple_DQP}).
\section*{Simulation results}
The proposed linearly adaptive loss function has been tested on a deep learning model based on the ResNet (Residual Network) architecture, a widely used deep learning model designed for tasks such as image classification \cite{he2016deep} and built around the concept of residual blocks or ``skip connections," which allow the network to learn identity functions and alleviate the problem of vanishing gradients in very deep networks. Our test model consists of 18 layers. The fundamental building unit of the ResNet architecture, residual block, contains two convolutional layers and a shortcut connection. If the input and output of the block have different dimensions, a convolutional layer is added to the shortcut to adjust the dimension. The ResNet starts with a convolutional layer followed by batch normalization. Then it constructs 4 layers, each containing a sequence of residual blocks. After the 4 layers, an average pooling layer and a fully connected layer are added for the final classification. 

This model is designed for the CIFAR-100 dataset, which consists of 100 classes of 32x32 color images. To better handle the images in this dataset, a per-pixel mean subtraction operation is performed as a preprocessing step. This operation helps to center the data and can improve the performance of the model. It measures the probability that the true class is among the top 5 classes predicted by the model.

Key hyperparameters and factors involved in the process are outlined as follows:
\begin{enumerate}
\item Optimizer: Stochastic Gradient Descent (SGD) is used as the optimization algorithm to minimize the loss function. The learning rate is set to 0.1, the momentum to 0.9, and weight decay (which helps in avoiding overfitting by acting as a regularization term) is set to 5e-4.

\item Learning Rate Scheduler: A StepLR scheduler is used which adjusts the learning rate after a fixed number of epochs. The learning rate is decayed by a factor of 0.1 every 50 epochs, which helps in converging to a solution more efficiently.

\item Training Epochs: The model is trained for 200 epochs. An epoch corresponds to one complete pass through the entire training dataset.

\item Batch Size: The batch size, which is the number of training examples utilized in one iteration, is set to 100 for the train and test loaders.

\item Data Augmentation: The training images undergo data augmentation processes which include random horizontal flip and random cropping. This increases the effective size of the training data and helps the model generalize better.

\item Normalization: Per-pixel mean subtraction is used as a preprocessing step to normalize the images. This can help speed up training and make the process less sensitive to the initial weights.
\end{enumerate}

\begin{figure}
\center
\includegraphics[scale=1]{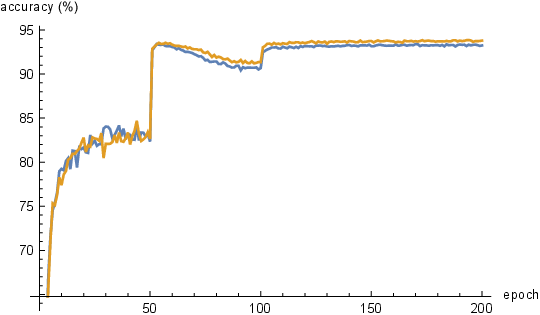}
\caption{Comparison of Test Data Accuracy between Cross Entropy and Linearly Adaptive Loss Functions. The graph illustrates the comparative performance of two distinct loss functions over the course of training. The upper line (orange) depicts the accuracy achieved using the Linearly Adaptive Loss function, while the lower line (blue) represents the results from the Cross Entropy Loss function. Each line is an average of five separate training iterations. This comparative analysis underscores the performance distinction between the two models throughout their training process.}
\label{fig:Adp}
\end{figure}

\begin{figure}
\center
\includegraphics[scale=1]{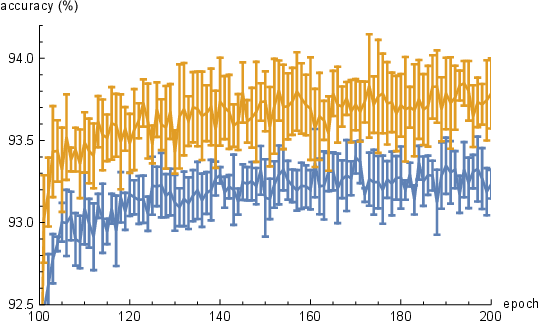}
\caption{Detailed View of Fig.~\ref{fig:Adp} Focusing on Epochs 100 through 200. This figure provides a zoomed-in analysis of the training process between epochs 100 and 200, highlighting the performance dynamics during this specific interval. The incorporated error bars illustrate the standard deviation at each individual epoch.}
\label{fig:focusImage}
\end{figure}

We have compared the performance of two different loss functions: Cross Entropy and Linearly Adaptive Losses, using the same dataset described above for a set number of training iterations. Fig.~\ref{fig:Adp} displays the comparative performance of these two loss functions over the course of the training process. 

We found that the Linearly Adaptive Loss function, represented by the orange (upper) line, consistently outperformed the Cross Entropy Loss function, as indicated by the blue (lower) line. Each line is an average of five separate training iterations, which were conducted to mitigate the impact of stochastic elements in the training process and offer a robust measure of each function's performance.

To examine the performance in greater detail, Fig.~\ref{fig:focusImage} provides a zoomed-in view of the training process, focusing on epochs 100 through 200. It offers a nuanced understanding of the performance dynamics during this critical phase of the training process. The error bars incorporated in this figure represent the standard deviation at each individual epoch, thereby providing insight into the variance in performance across the different training iterations. Table.~\ref{table:comp} shows the values of the top-5 error rates for each trial under the Cross Entropy and Linearly Adaptive Loss functions. These error rates have been computed by averaging the results from epochs 190 to 200, giving a more stable view of the model's performance. The table reveals that the Linearly Adaptive Loss function consistently outperforms Cross Entropy, attaining a lower mean top-5 error rate and provides standard deviations for these error rates, allowing for a consideration of consistency in addition to raw performance.

\begin{table}[ht]
\begin{tabular}{ccc}
\hline
loss function    & top-5 error (\%) & trial no.  \\
\hline \hline
cross entropy      & $6.7 \pm 0.1$        & mean and std.      \\
                           & 6.8 &   trial 1 \\
                           & 6.6 &  trial 2\\
                           & 6.8 & trial 3\\
                           & 6.6 & trial 4\\
                           & 6.7 & trial 5\\
linearly adaptive loss      &$\textbf{6.2} \pm 0.15$        &  mean and std.      \\
                           & 6.2 & trial 1\\
                           & 6.2 & trial 2\\
                           & 6.1 & trial 3\\
                           & 6.5 & trial 4\\
                           & 6.2 & trial 5\\
                           
\bottomrule

\hline
\end{tabular}
\caption{Comparison of test accuracy for the CIFAR-100 dataset when employing Cross Entropy and Linearly Adaptive Loss functions. Mean accuracy and corresponding standard deviation were derived from five trials for each loss function, specifically using data from epochs 190 to 200. Linearly Adaptive Loss demonstrated superior performance with a lower top-5 error rate.}
\label{table:comp}
\end{table}

\section*{Computation cost}
The additional computational cost of our proposed loss function compared to the standard cross entropy loss is that it requires two extra operations which are a subtraction and a multiplication. These operations do not significantly increase the computational burden.

\section*{Conclusion}
We introduced and analyzed the performance of a linearly adaptive cross entropy loss function, constructed to enhance the optimization process for classification tasks. The key feature of the proposed function is the additional term that depends on the predicted probability of the true class. This aspect allows the model to focus more on correctly classifying the true class, thereby potentially improving overall classification performance. In the application to a ResNet-based model trained on the CIFAR-100 dataset, the linearly adaptive loss function exhibited superior performance compared to the conventional cross entropy loss function, in terms of accuracy. The findings from our empirical analysis suggest that the linearly adaptive cross entropy loss function could provide an effective alternative for loss function optimization in machine learning models. Our findings underscore the potential of the proposed linearly adaptive cross entropy loss function as a more accurate alternative for classification tasks. However, this is only the first step in exploiting the potential of this new loss function.

Potential future work could involve a detailed theoretical analysis of the convergence properties of our linearly adaptive loss function, which might provide insights into why it performs better than traditional cross entropy loss in certain scenarios. In addition, the impact of this loss function on the robustness of models to adversarial attacks \cite{szegedy2013intriguing}, which are small perturbations to the input data that can significantly degrade the model's performance, is an interesting area for exploration \cite{goodfellow2014explaining}. Lastly, exploring the potential of the linearly adaptive cross entropy loss function for multilabel classification tasks could also yield promising results, given its inherent design of focusing more on the true class \cite{zhang2013review}.


\section*{Additional Information -- competing interests}
The author(s) declare no competing interests.

\section*{Data availiability}
The datasets used and/or analysed during the current study available from the corresponding author on reasonable request.

\section*{Funding}
This work was partially supported by the KIST Institutional Program.
\section*{Conflicts of interest}
Not applicable.
\section*{Ethics approval}
Not applicable.
\section*{Consent to participate}
Not applicable.
\section*{Consent for publication}
Not applicable.
\section*{Availability of data and material}
Not applicable.
\section*{Code availability}
Not applicable.
\section*{Authors' contributions}
This work was done by Jae Wan Shim.

\end{document}